\documentclass[runningheads]{llncs}
\usepackage[T1]{fontenc}
\usepackage{lmodern}
\usepackage{microtype}
\let\subparagraph\paragraph 
\usepackage{titlesec}
\titlespacing*{\section}{0pt}{6pt plus 1pt minus 1pt}{4pt plus 1pt minus 1pt}
\titlespacing*{\subsection}{0pt}{5pt plus 1pt minus 1pt}{3pt plus 1pt minus 1pt}
\titlespacing*{\subsubsection}{0pt}{4pt plus 1pt minus 1pt}{2pt plus 1pt minus 1pt}
\let\oldthebibliography\thebibliography
\renewcommand{\thebibliography}[1]{\oldthebibliography{#1}\setlength{\itemsep}{0pt}\setlength{\parskip}{0pt}\setlength{\parsep}{0pt}}
\usepackage{comment}
\usepackage{float}
\usepackage[dvipsnames]{xcolor}
\usepackage{cite}
\usepackage{booktabs}
\usepackage{setspace}

\usepackage{wrapfig}
\usepackage{graphicx}
\usepackage{algorithm}
\usepackage{amsmath}
\usepackage{algpseudocode}
\usepackage{marvosym} 
\usepackage{color}
\usepackage{hyperref}
\hypersetup{hidelinks}
\usepackage{listings}
\usepackage{tabularx}
\microtypesetup{expansion=true, protrusion=true, factor=1100}
\usepackage{array}
 \newcolumntype{Y}{>{\centering\arraybackslash}X}
\usepackage[normalem]{ulem} 
\newcommand{\MS}[1]{#1}

\begin{document}
\title{A Negotiation-Aware Hybrid LLM Framework for Intent Interpretation and Service Discovery}
\titlerunning{A Negotiation-Aware Hybrid LLM Framework}
\author{
Wided Meflah\textsuperscript{(\Letter)}\orcidID{0009-0007-9794-4356} \and
Hayet Brabra\orcidID{0000-0001-7484-2268} \and
Walid Gaaloul\orcidID{0000-0003-0451-532X}
}
\authorrunning{W. Meflah et al.}
%
\institute{SAMOVAR, Telecom SudParis, Institut Polytechnique de Paris, Palaiseau, France\\
\email{\{wided.meflah,hayet.brabra,walid.gaaloul\}@telecom-sudparis.eu}}
\maketitle              
\begin{abstract}


Increasingly, users express their needs as natural-language intents rather than structured requests. These intents may be incomplete, high-level, or internally conflicting. A conflicting intent either violates fundamental properties of the cloud paradigm or states mutually inconsistent requirements. Classical composition approaches are designed for structured, low-level, and complete requests. They cannot interpret such intents, discover suitable services from provider catalogs, or negotiate conflicting requirements with users. To address this gap, we propose a negotiation-aware framework for intent interpretation and service discovery. The framework first refines the natural-language intent and dynamically detects conflicts. It then structures the intent into a provider-independent TOSCA template. For each specification, it retrieves candidate service offers through deterministic matching grounded in cloud provider offerings. Some conflicts arise during interpretation. Other requirements prove unsatisfiable during discovery. Both are resolved through iterative natural-language negotiation with the user. We evaluate the framework on 70 diverse intents against single-shot and enhanced-prompting baselines. It improves recall and F1 for both conflict detection and service discovery. It also eliminates hallucinated offers and resolves conflicts at a cost proportional to their number.
\keywords{Cloud Service Composition \and Intent Interpretation \and Service Discovery \and Large Language Models  \and TOSCA.}
\end{abstract}
%
%
%
\section{Introduction}

Modern cloud-native applications typically combine several autonomous services into a higher-value solution, a process known as service composition (SC)~\cite{syu2023gap,lamine2018formal}. This process begins with request interpretation, which maps the user's request into a structured, provider-independent set of required services, followed by service discovery, which retrieves, from provider catalogs, the provider offers that fulfill the request, followed by service selection, which chooses the most appropriate offers among the discovered candidates, and finally code generation, which produces the composite service's executable code~\cite{meflah2025natural,sheng2014web,brabra2019model}. Classical SC approaches are designed for structured, low-level, and complete requests; however, non-expert users express their needs as free-form natural-language (NL) intents~\cite{meflah2025natural,meflahhybrid,brabra2026beyond}, which we refer to as NL requests throughout this paper. Such NL requests are typically high-level, incomplete, and potentially conflicting. Consider an NL request asking to "store large volumes of data, and connect it locally to another component hosted in a different region." This request already exposes the difficulties of free-form requests. It is high-level: "store data" must be mapped to a service, e.g., a MySQL database. It is incomplete as it does not mention the compute resource (e.g., a virtual machine) needed to host that database. Finally, the request is conflicting. In the example, a local (L2) network connection between components located in different regions violates a fundamental limitation of the cloud paradigm, regardless of the provider chosen~\cite{jerbi2023request}, since an L2 link needs a single broadcast domain while regions are interconnected only at L3 and cannot share one natively. We call this a cloud-paradigm conflict type. A different conflict type may arise as an inconsistency between two of the user's own requirements rather than a cloud-paradigm limitation, which we call a user-requirement conflict; for instance, when a component is required to scale automatically while also being pinned to a fixed IP address. Finally, even a request free of both conflict types may still prove unsatisfiable in practice, e.g., if no provider catalog actually offers the requested database engine in the specified region; we call this a discovery conflict, since it only surfaces during service discovery, once matching against real offerings is attempted. The abstraction, incompleteness, and conflicts that characterize such NL requests expose specific limitations in existing composition approaches. Methods based on NLP and supervised learning~\cite{zhao2016mining,adala2011framework} rely on large labeled datasets, which restricts their applicability to free-form, previously unseen NL requests. LLM-based approaches~\cite{huang2023let,song2023restgpt} are typically prompted directly on the request; however, prior work~\cite{meflah2025natural} shows that this strategy is insufficient, as LLMs omit high-level and implicit requirements, and hallucinate non-existent services. Embedding the full catalog in the prompt~\cite{song2023restgpt} mitigates hallucination but does not scale, since the token cost grows with the size of the service catalog. Retrieval-augmented (RAG) approaches~\cite{bianchini2024leveraging} bound this cost, yet they return services that are textually similar to the request rather than provably compliant with its exact requirement values, a limitation we confirm empirically in Sect.~\ref{sec:evaluation}.\looseness=-1

Overcoming these limitations requires two fundamentally different capabilities. Interpreting a high-level, incomplete, or conflicting request requires contextual reasoning, a generative task best suited to LLMs, which interpret the request into a structured, provider-independent request. Discovery, in contrast, requires an exact, binary decision over well-defined requirement values, and is therefore best suited to deterministic matching, which guarantees that returned offers are provably compliant and free of hallucination. When conflicts arise, rather than rejecting the request or returning the nearest approximate match, our solution proposes relaxations of the conflicting constraints and presents these options to the user, who selects the best fit through NL interaction. For the L2 connection above, for example, it might propose replacing it with an L3 connection or co-locating the two components in the same region. We therefore propose a negotiation-aware framework that combines three complementary mechanisms. First, an enhanced LLM refines the request, and a specialized LLM structures it as a TOSCA~\cite{oasis2013orchestration} template, a provider-independent standard for describing cloud applications, well suited to the interpretation step. Second, a deterministic algorithm performs discovery against the provider catalogs using this template. Third, a shared LLM-based mechanism resolves conflicts through negotiation with the user, both during interpretation and during discovery. Overall, we extend the two traditional SC steps, interpretation and discovery. We enrich interpretation with a refinement that reformulates high-level requests and completes incomplete ones, and we enrich both steps with conflict detection, which identifies unsatisfiable sets of constraints, conflict relaxation, which proposes minimal adjustments that restore satisfiability, and NL negotiation, through which the user chooses among these adjustments. Our contributions are as follows. \textit{(i)} We extend request interpretation with refinement and dynamic detection of conflicts over both functional and non-functional requirements, covering cloud-paradigm and user-requirement conflict types, on a provider-independent structured representation. \textit{(ii)} We address discovery with an algorithm that performs binary compliance checks against the actual catalogs, eliminating hallucinated offers and explicitly exposing unsatisfiable requirements. \textit{(iii)} We design a single conflict relaxation and negotiation mechanism, reused across both the interpretation and discovery phases, that relaxes and negotiates every detected conflict or unsatisfiable requirement with the user. \textit{(iv)} We evaluate the framework on 70 diverse NL requests against single-shot and enhanced-prompting baselines, showing improved recall and F1 for both conflict detection and discovery, zero hallucination, and relaxation at a cost proportional to the number of conflicts.\looseness=-1


\section{Related work}\label{sec:sota}

In this section, we review prior work along the two SC steps targeted by our approach: request interpretation and service discovery.

Regarding request interpretation, automated service composition has traditionally operated on structured or semi-structured requests~\cite{syu2023gap,lamine2018formal,brabra2023uncovering}. Such approaches assume low-level, complete requests, and do not consider free-form natural-language requests increasingly issued by non-expert users, which may be high-level, incomplete, or conflicting~\cite{meflah2025natural}. To handle natural-language requests, a first line of work applies traditional NLP and supervised learning~\cite{zhao2016mining,adala2011framework} to convert them into structured service requests. However, these methods require large labeled datasets, which limits their applicability to free-form, unseen requests. More recently, LLM-based approaches~\cite{huang2023let,song2023restgpt} interpret requests without task-specific training but, prompted directly on the request in a zero-shot manner, miss high-level, implicit requirements, do not consider conflicts, and hallucinate non-existent services. A recent multi-step LLM framework~\cite{meflah2025natural} addresses part of this by refining requests of varying abstraction and completeness before producing a structured output; however, it covers only functional needs and neither detects conflicts nor supports their relaxation and negotiation with the user. Jerbi et al.~\cite{jerbi2023request} represent requests and offers through a shared capability model and, when a request cannot be matched, identify the responsible provider constraints and recommend relaxation plans that rewrite the request accordingly. However, these relaxations are recommended in a single shot, with no user in the loop, since the approach operates on structured requests without NL interaction; moreover, its supported constraints are fixed by design, so extending it to new conflict types would require redefining the mechanism rather than handling them dynamically. In contrast, our work interprets high-level, incomplete, and potentially conflicting NL requests, covers both functional and non-functional needs, and resolves detected conflicts through iterative NL negotiation with the user.

Regarding service discovery, one established line of work formulates it as ontology-based semantic matchmaking~\cite{alkalbani2021cloud,gu2024semantically}, matching requests and offers against shared ontologies by computing degrees of correspondence over their inputs and outputs. However, such matchmaking yields a graded similarity score, whereas cloud service discovery requires returning only the offers that provably satisfy the requested values. A second line models discovery as a supervised learning problem~\cite{tang2024mashup,yuan2024intelligent}, mapping need descriptions to services through classical or transformer-based classifiers. However, these methods require a large annotated corpus, which is costly to build. More recently, LLM-based approaches have been applied to service discovery. A first group prompts the model directly on the request~\cite{huang2023let,peng2025llmsrec}, which frequently proposes non-existent services. To ground the answer, other works embed the service descriptions within the prompt~\cite{song2023restgpt}, or distribute them across one specialized agent per service category~\cite{liao2024hmsc,li2025llm}; both remain generation-based, so hallucination persists, and the latter further multiplies the number of LLM calls. Fine-tuning an LLM for discovery~\cite{qin2025harnessing} is likewise ill-suited: it treats an information-retrieval task as a generative one and memorizes the catalog in its weights, so any update requires retraining. Another line of work uses retrieval-augmented generation~\cite{bianchini2024leveraging}, retrieving candidate offers by embedding similarity and letting the LLM verify them against the request. This keeps the catalog updatable and grounds the answer in retrieved offers. However, retrieval returns offers textually close to the request rather than ones that provably satisfy it, becomes ineffective when a service is specified by name, and still returns the least-dissimilar candidates when none complies. We confirm this limitation empirically in Sect.~\ref{sec:evaluation}. In contrast, our work performs discovery with a deterministic matching algorithm that grounds every candidate in the actual catalogs: it returns exactly the offers satisfying each requirement, remains effective when a service is specified by name, and is free of hallucination. When no offer satisfies a requirement, it identifies the unsatisfiable requirements rather than falling back on the least-dissimilar candidate, and resolves them, relying on an LLM only at this step, through iterative NL negotiation with the user.\looseness=-1

\section{Preliminaries}\label{sec:prelem}

Between interpretation and discovery, our framework relies on an intermediate representation of the user request. To let discovery match this request against offerings from any catalog, the representation must be independent of any provider or technology, while capturing the services to compose together with their parameters, dependencies, and non-functional requirements. Grounding it in a community-adopted standard further keeps the output reusable. We use TOSCA, a widely used OASIS standard that meets these needs.
TOSCA~\cite{oasis2013orchestration} defines a data model to describe, deploy, and manage cloud applications independently from providers and technologies. A cloud application is described by a topology template of node templates representing its components (e.g., Compute, Database, or Block Storage). Each node template has a node type fixing its properties (configurable parameters, e.g., a Database's name, user, and password), its capabilities (functionalities offered to other nodes, e.g., a Compute node's host capability with CPU, RAM, etc.), and its requirements (dependencies on capabilities, e.g., a DBMS requiring a host). A topology template also defines policies expressing non-functional requirements over target nodes, each with a policy type fixing its semantics; e.g., a Placement policy constrains deployment location (region or availability zone) and a Performance policy declares latency or response-time requirements, while custom types capture others such as Cost. We represent the request interpretation output as such a TOSCA template. Any equivalent provider-independent abstraction language could nonetheless play the same role without affecting the rest of the framework.

\section{Approach Overview}
\label{sec:Overview}
We propose a framework that takes a NL request and produces a set of candidate provider offers fulfilling it, automating the first two steps of cloud service composition, namely request interpretation and service discovery.\begin{figure}[ht]
\centering
\includegraphics[width=0.56\textwidth]{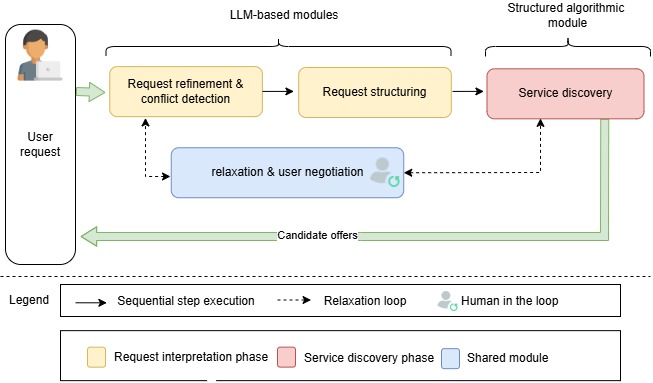}
\caption{Overview of our framework}
\label{fig:overview}
\end{figure} As shown in
Figure~\ref{fig:overview}, it is organized into these two phases, connected by a shared conflict relaxation and negotiation module invoked whenever a set of requirements cannot be satisfied. Request interpretation turns the user's request, which may vary in abstraction and completeness and may contain conflicts, into a structured, provider-independent TOSCA template, refining high-level needs into services, completing missing elements, and detecting conflicts. Service discovery then matches each required service against the provider catalogs with a deterministic algorithm and returns the offers satisfying its functional and non-functional requirements. Whenever a conflict is detected during interpretation or a requirement cannot be satisfied during discovery, the shared module proposes, in natural language, one or more relaxation options that would make the request satisfiable. The user intervenes only at this point, and only to select, for each conflict or unsatisfiable requirement, the option they prefer. The framework then merges the selected relaxations and retries matching, so the user never reformulates the request, validates offers, or edits the architecture. This loop continues until the request becomes conflict-free and satisfiable. The two phases are
detailed in Sections~\ref{sec:interpretation} and~\ref{sec:discovery}.
\section{Request Interpretation}
\label{sec:interpretation}

This phase interprets the user's NL request, which may be expressed at varying
levels of abstraction and completeness and may contain conflicts. It comprises
three modules. The first is dedicated to request refinement and conflict
detection. The second handles the relaxation of conflicts, if any, and their
negotiation with the user; it is shared with the service discovery phase, where
it is reused to handle requirements that no available offer can satisfy, so we
describe it here and refer back to it in that phase. The third generates the structured output as a TOSCA template, the provider-independent intermediate model motivated in Section~\ref{sec:prelem}, which lets the deterministic discovery phase match each service against any provider catalog.
\subsection{Request Refinement and Conflict Detection}
\label{subsec:refinement}
\begin{figure}[ht]
    \centering
    \includegraphics[width=0.61\textwidth]{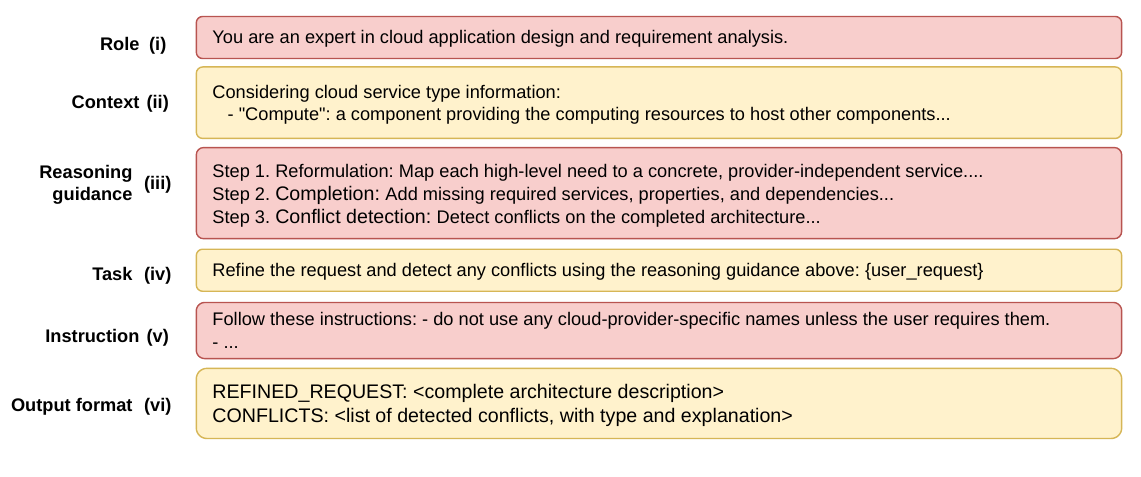}
    \caption{Excerpt of the request refinement and conflict detection prompt template.}
    \label{fig:prompt-template}
\end{figure}
The first module of the request interpretation phase carries out two
complementary tasks: it refines the NL request into a complete,
low-level request, and it detects the conflicts that this request
may contain. We perform both in a single LLM call, since conflict detection operates on the refined architecture and shares its context, avoiding redundant calls and token cost. This call uses a prompt following the structure shared by every LLM call in our framework with variations only in their content (Fig.~\ref{fig:prompt-template}): the Role~(i) casts the model as an expert in cloud application design and requirement analysis; the Context~(ii) provides the catalog of TOSCA provider-independent service types with their properties, capabilities, and requirements; the Reasoning guidance~(iii) is a chain-of-thought of three steps: reformulation, completion, and conflict detection, detailed below, which improves output reliability; the Task and Specific Instructions~(iv,~v) state the task and its detailed expectations; and the Output format~(vi) specifies the structured result, namely the refined request with the detected conflicts.

\textit{Reformulation.} When the request expresses business needs, the module first reformulates it into provider-independent services, for instance, ``store data'' is mapped to a database, while vague sizing descriptions are
turned into measurable specifications \MS{whose values are set by the LLM from its knowledge of typical cloud configurations} (e.g., ``large memory'' $\rightarrow$
16~GB of RAM). Non-functional needs expressed informally are likewise turned into explicit, measurable requirements, for instance, "cheap" into a cost limit. Already service-oriented requests bypass this step.

\textit{Completion.} A request considered incomplete is then completed by
adding the elements required to make the architecture deployable. To guide this
process and limit hallucinations, the prompt is statically enriched with
structured knowledge derived from the TOSCA
specification\footnote{\url{https://docs.oasis-open.org/tosca/TOSCA-Simple-Profile-YAML/v1.3/TOSCA-Simple-Profile-YAML-v1.3.html}},
namely the normative node types together with their properties, their
capabilities and requirements. We rely on this normative knowledge rather than
on the model's parametric knowledge of TOSCA, which is often incomplete or
outdated, so that every added component remains compliant with the
underlying model. Consider, for instance, a request that mentions only a MySQL
database. The module infers that the database must be managed by a DBMS, which
in turn must be hosted on a compute resource. It therefore adds the
corresponding DBMS and virtual machine, assigns their required properties
(e.g., the DBMS version and the machine's CPU, memory, and disk size). Complete requests bypass this step.

\textit{Conflict detection.} Once the request is complete, this module
analyzes it to detect conflicts, i.e., sets of constraints that cannot be
satisfied simultaneously. We perform this detection with an LLM rather than a hardwired procedure
so that conflicts can be handled dynamically, covering varied and previously
unseen conflicts without redefining the mechanism for each new case. We distinguish two types. Cloud-paradigm conflicts arise between the
request and the inherent limitations of the cloud
paradigm~\cite{wright2012constraints,jerbi2023request}, which hold across all
providers, for instance, requesting an L2 connection between two nodes located in different regions.
User-requirement conflicts arise among the user's own constraints~\cite{chen2012systematic},
when two or more explicitly stated requirements are mutually inconsistent, for
example, contradictory co-location constraints, where transitive co-location
requirements force the same component into two incompatible locations. We detect conflicts on the completed request rather than the raw request, since some surface only once implicit services and dependencies are explicit. This module thus outputs the refined request and its detected conflicts, if any: when present, both are forwarded to the relaxation and negotiation module; otherwise the refined request proceeds directly to structuring.
\subsection{Conflict Relaxation and User Negotiation}
\label{subsec:relaxation}

The second module is triggered when the previous module reports one or
more conflicts. As noted above, the same module is reused in the service
discovery phase to propose concrete relaxation options that make
the conflicting constraints satisfiable again, and to let the user decide,
through an interactive negotiation, which of these options to apply. We delegate this module to a specialized LLM because relaxation requires reasoning about the user's intent and expressing the proposed alternatives in natural language, so that they can be understood and acted upon. Its prompt follows the shared template of Figure~\ref{fig:prompt-template}, giving the model the request and the reported conflicts as context and asking it to return, for each conflict, a set of relaxation options in natural language. For each conflict independently, the model derives a minimal set of resolutions that removes the incompatibility, ensuring that a change addressing one conflict does not interfere with the others.\looseness=-1

\textit{Relaxation.} For every detected conflict, the module generates at least
one relaxation option, favoring the smallest change that restores
satisfiability to keep the relaxed request as close
as possible to the user's original request. The strategy depends on the conflict
type. For cloud-paradigm conflicts, the request is aligned with what the cloud paradigm
allows; for instance, for an L2 connection between nodes in different regions, propose an L3 connection instead or place the nodes in the same region. For user-requirement conflicts, one
or more user-stated constraints are loosened so that the remaining ones become
jointly satisfiable; for instance, when transitive co-location requirements force the same component into two incompatible locations, the module proposes dropping the weakest co-location constraint so that a consistent placement becomes possible. \MS{These strategies guide the LLM rather than forming a fixed rule set: the concrete options are generated for each detected conflict.}

\textit{Negotiation.} The generated options are presented to the user, who remains in the loop and selects one option per conflict through a turn-based natural-language exchange with the framework. \MS{All options of a turn are presented at once, so the user resolves every pending conflict in a single interaction.} The selected relaxations are then
merged with the original request, and the resulting augmented request is sent
back to the refinement and conflict-detection module, which re-interprets it if
necessary and re-checks it for new conflicts. The loop between the two modules
continues until no conflict remains, at which point the conflict-free request
proceeds to the request structuring module. 


\subsection{Request Structuring}
\label{subsec:structuring}

The third module produces the final structured output of the interpretation
phase, namely a provider-independent TOSCA template that captures the refined,
conflict-free request. We introduce a dedicated structuring step, rather than
letting the model emit the template directly from the request, so that
generation is guided by a known schema and remains compliant with the TOSCA
specification. The template is generated by a single LLM call whose prompt is statically
enriched with the same TOSCA-derived knowledge as above, in order to limit the
hallucinations commonly observed when LLMs produce TOSCA directly, such as generating
non-existent node types or invalid property names.
Guided by this knowledge, the module maps each service in the request to a
TOSCA node template, fills
the corresponding TOSCA properties with the parameter values stated in the
request, expresses each dependency by matching the requirement of one node with
the capability of another, and fills TOSCA policies with the non-functional
requirements of the request. The resulting template is the output
specified for the request interpretation phase and serves as the input to the
second phase. This step runs automatically with no human validation; across the whole pipeline, as stated in Section~\ref{sec:Overview}, the user intervenes only during the negotiation (Section~\ref{subsec:relaxation}).\looseness=-1

\section{Service Discovery}\label{sec:discovery}

This phase retrieves, from the service catalogs, the provider offers that fulfill
each provider-independent service of the request produced by the previous phase.
Matching is performed by a deterministic algorithm rather than an LLM, since cloud service discovery is fundamentally an information-retrieval task rather than a generation one: it requires an exact, binary determination of whether an offer satisfies the structured properties and policies of a requirement, rather than the graded measure of correspondence that semantic, LLM-based matching computes. Because it is grounded in the actual offerings, the algorithm is immune to hallucination, returns the complete set of satisfying candidates, and handles service-name-based requirements that defeat semantic matching. Involving no LLM, it incurs neither token cost nor multiple-call overhead, and scales to large catalogs that stay updatable without retraining.
The
LLM is therefore confined to the relaxation step: when no offer is found for one or
more services, the unsatisfiable requirements are handed back to the relaxation and
negotiation module shared with the interpretation phase (Section~\ref{sec:interpretation}). The user is the only human in the loop and plays the same role as in that phase, selecting for each unsatisfiable requirement one of the relaxation options the module proposes in natural language\MS{, phrased in plain terms rather than as TOSCA properties (e.g., ``use PostgreSQL in the requested region'')}. The adjusted request is then re-matched, and this
detection--relaxation--matching loop repeats until every service obtains at least
one candidate offer or the user ends the negotiation. Operating on the TOSCA template produced by the interpretation phase, the algorithm
groups together nodes with identical requirements and matches each group against
the catalogs, so that identical requirements are matched only once. We maintain
one catalog per service type, namely compute, load balancer, database, network,
object storage, and block storage, modeled as TOSCA node types and covering the
core service types needed to compose typical cloud applications. Separating them by
type restricts the search for each requirement to the offers of the relevant type
and lets a type be updated independently of the others. Each catalog stores the
offers of the supported providers, where an offer is a concrete, regionalized
service described by its functional specifications and its non-functional attributes.

Having grouped nodes with identical requirements (line~1), the algorithm iterates over the resulting groups (line~2) and proceeds in three steps, summarized in Algorithm~\ref{alg:service-matching}. \textit{Requirement extraction} (line~3) derives, for each group, the constraints to satisfy from its type, properties, capabilities, and policies, where the type gives the service type to search for, the properties and capabilities give the functional requirements (e.g., CPUs, memory, or database engine and version), and the policies give the non-functional ones (regions, availability zones, providers, minimum availability, and cost limit). \textit{Candidate filtering} (line~4) then selects, among the offers of that service type, those satisfying every extracted requirement, each treated as an independent criterion the offer must pass. \MS{Compliance is thus verified per requirement, whereas cross-service compatibility (e.g., joint placement) pertains to service selection, which lies outside the scope of this work.} \textit{Candidate set construction} (line~5) forms the group's candidate set from the offers passing all criteria. If a group yields none, the algorithm does not fall back on the closest offer; it instead reports the criteria responsible for the failure as unsatisfiable requirements (line~5), such as a required region in which no offer provides the requested capacity, or an availability threshold that no allowed provider reaches, which are handed to the relaxation and negotiation module as described above. Otherwise, the candidate offers retrieved for all groups form the output of this phase (line~6).
The algorithm runs in $O(|G| \cdot m \cdot k)$ time, where $|G|$ is the
number of requirement groups ($|G| \leq |T.\mathit{nodes}|$ thanks to
grouping), $m$ the size of the largest catalog, and $k$ the bounded number
of criteria per requirement. It thus scales linearly with catalog size,
confirming its suitability for large, evolving catalogs.
\begin{algorithm}[H]
\caption{Service Matching}
\label{alg:service-matching}
\setstretch{0.85}
\footnotesize
\begin{algorithmic}[1]
\Require TOSCA template $T$, catalogs $\mathcal{R}$ (one per service type)
\Ensure Candidate sets $\mathcal{C}$, or unsatisfiable requirements $U$
\State $\mathcal{C} \gets \emptyset$;\ $U \gets \emptyset$;\ $G \gets \textsc{GroupByRequirements}(T.\text{nodes})$
\For{$g \in G$}
    \State $\tau \gets g.\text{type}$;\ $F \gets g.\text{properties} \cup g.\text{capabilities}$;\ $N \gets g.\text{policies}$
    \State $\mathit{cand} \gets \{\, o \in \mathcal{R}[\tau] \mid \textsc{Satisfies}(o,F) \wedge \textsc{Satisfies}(o,N) \,\}$
    \State \textbf{if} $\mathit{cand} = \emptyset$ \textbf{then} $U \gets U \cup \textsc{FailedCriteria}(g, \mathcal{R}[\tau])$ \textbf{else} $\mathcal{C}[g] \gets \mathit{cand}$
\EndFor
\Return $U$ if $U \neq \emptyset$ \textbf{else} $\mathcal{C}$
\end{algorithmic}
\end{algorithm}

\section{Evaluation}
\label{sec:evaluation}
\subsection{Experimental Setup}
\subsubsection{Dataset} We evaluate our framework on 70 natural-language requests. Twenty-four come from a public benchmark for natural-language service
  composition~\cite{meflahhybrid}, which supplies three request forms (complete low-level, incomplete low-level, and business-oriented) and, being conflict-free, lets us evaluate on requests without conflicts. The remaining 46 are built around conflicts of three types. For every conflict, we write the request in the three mentioned request forms, so that detection is tested across varying completeness and abstraction levels, and we add an associated conflict-free near-miss request of identical structure to guard against false positives; for each type, one or two additional requests combine a pair of conflicts to confirm that co-occurring conflicts are all detected. The first type concerns cloud-paradigm violations that no provider can satisfy; drawing on the literature, we cover four of them: an L2 connection between different regions~\cite{jerbi2023request}, a block-storage volume shared across VMs~\cite{oasis2013orchestration}, in-place modification in object storage~\cite{mansouri2017data}, and geographic separation incompatible with low latency~\cite{wright2012constraints}. This yields 18 requests. The second type concerns mutual inconsistencies among the user's own requirements; we cover three of them: contradictory co-location constraints~\cite{chen2012systematic}, contradictory co-location combined with anti-co-location constraints~\cite{chen2012systematic}, and a component that is both scalable and bound to a fixed IP~\cite{oasis2013orchestration}. This yields 13 requests. \MS{Since the prompts list no concrete conflict, these seven interpretation conflicts are unseen by the model.} The third type, discovery conflicts, is of a different nature: reasonable requirements that none of the offers in our catalogs satisfies. We study seven, all verified against our catalogs: two where two non-functional requirements exclude one another (performance vs.\ cost, availability vs.\ cost), and five where a requested feature is offered nowhere within its scope, namely a database engine and a block-storage capacity unavailable in the requested region, a service in a region covered by no offer, and a load-balancer scope and a network protocol unavailable from the requested provider. This yields 15 requests.

Our dataset is structured into seven columns: (1) an identifier; (2) the NL request; (3) its request form (complete low-level, incomplete low-level, or business-oriented); (4) its conflict type (conflict-free, cloud-paradigm, user-requirement, or discovery conflict); (5) the ground-truth conflicts it carries; (6) the admissible relaxations for these conflicts\MS{, derived from the cited literature and manually annotated by the authors}; and (7) a near-miss flag marking the conflict-free requests built to mirror a conflicting one. This ground truth is fixed for conflicts and relaxations. For the offers it is instead established per execution, since the final candidate set depends on the relaxations the user applies during negotiation.
\subsubsection{Framework implementation}

We implemented our proposed framework (Fig.~\ref{fig:overview}) in Python, using
LangChain\footnote{\url{https://www.langchain.com/}} to orchestrate the interactions
with the LLMs across the different modules. During development we evaluated several models (LLaMA~3.3~70B, GPT-4o~Mini, gpt-oss-120b, open-mistral-7b, mistral-large-2512) and selected mistral-large-2512 for all three LLM-based modules as the most accurate; we set the temperature to 0.1 for output stability.
The service discovery algorithm operates on the provider catalogs, which we
maintain as one table per service type: compute, network, database, load balancer,
object storage, and block storage. These six types correspond to normative cloud
node types of the TOSCA specification.
Each offer has attributes shared across types (provider, region, sla, price) and type-specific ones (e.g., cpu and memory for compute, engine for database); a separate taxonomy table resolves logical locations into each provider's regions and availability zones. Overall, the catalogs gather 209 concrete, regionalized offers from three major providers (AWS, Azure, GCP), exercising the matching algorithm on diverse, real services. All experiments run in fully automatic mode: rather than negotiating with the user, the system retains, for each conflict or unsatisfiable requirement, the minimal relaxation proposed by the module, as it stays closest to the original request. \MS{This setting also ensures a fair comparison with the baselines, which involve no user.} The dataset, implementation code, and evaluation results are
available on GitHub\footnote{\MS{\url{https://github.com/widedMeflah/ICSOC2026}}}.
\subsubsection{Baselines}

\MS{The dedicated approaches of Sect.~\ref{sec:sota} are not directly comparable: \cite{meflah2025natural} covers neither non-functional requirements nor conflicts, and \cite{jerbi2023request} requires structured requests.} We compare our framework against two families of baselines, all given the same provider catalogs used by our framework. The first is a single-shot baseline, prompted once to refine the request, detect conflicts and propose relaxation plans, and discover a set of candidate offers that fulfill it. We run it with four models spanning proprietary and open systems, namely Claude Sonnet~4.6, Gemini~2.5~Flash, GPT-OSS-120B, and Mistral-Large-2512, the last being the model used within our framework, in order to assess whether the bare model, without our framework, suffices. A comparison against this single-shot baseline alone would not be a fair evaluation, as it pits a generic prompting scheme against our task-tailored pipeline. To isolate this, we compare our approach against a second family of baselines that keep the model fixed
 (Mistral-Large-2512, as used in our framework) and vary only the prompting strategy. Each strategy contributes one or more generic capability that our framework also relies on. We consider four strategies. Chain-of-Thought prompts the model to reason through our framework's steps (refinement, conflict detection, relaxation, TOSCA generation, and discovery). Self-Refine~\cite{madaan2023selfrefine} adds iteration, critiquing and revising an initial interpretation and candidate set through self-feedback until it stabilizes. RAG adds retrieval grounding\MS{, following the principle of~\cite{bianchini2024leveraging}}, injecting into the prompt the offers most similar to each requirement by embedding similarity and letting the model select among them. ReAct~\cite{yao2023react} adds an agentic reasoning--acting scheme, interleaving reasoning with actions that inspect the catalogs until it produces the final candidate set.\looseness=-1

\subsection{Evaluation Methodology}
We evaluate our framework along four complementary dimensions: the quality of conflict detection during interpretation and discovery, the quality of the relaxation options proposed for these conflicts, the quality of service discovery, and the computational cost incurred by the framework.
\subsubsection{Quality evaluation} We assess conflict detection, relaxation, and service discovery. For \textit{conflict detection}, since all baselines are prompted to detect conflicts, we use Precision, Recall, and F1-score to measure whether the reported conflicts correspond to ground-truth ones rather than spurious ones, whether all ground-truth conflicts are found, and to balance both. For \textit{relaxation}, we assess, for the detected conflicts, the quality of the proposed options using Recall@k, which measures whether at least one admissible relaxation recorded in the ground truth appears among the k options proposed for a conflict; we compute it for k = 5 so as to cover the full set of alternatives a method typically generates for a given conflict. For \textit{service discovery}, we evaluate the candidate offers each method returns from the same provider catalogs, using Precision, Recall, and F1-score, as above, and additionally a hallucination rate measuring the proportion of returned offers that are non-existent in the catalogs. 

\subsubsection{Computational cost}
We evaluate the computational cost incurred by our framework compared to the baselines. For each request, we measure the number of input and output tokens consumed, and the end-to-end latency, since these directly reflect the monetary and time costs of processing a request in practice. We report these measures separately for conflict-free requests, requests with one conflict, and requests with more than one conflict, as the number of conflicts a request contains directly affects the number of relaxation and re-matching iterations it triggers, and therefore the associated computational cost.

\subsection{Results and discussion}

\subsubsection{Quality analysis.}
Table~\ref{tab:results} reports the quality results.
For conflict detection, our framework attains the highest recall
(89.3\%) and F1 (79.0\%) while preserving a competitive precision
(70.8\%), owing to the reformulation and completion it performs, which
render conflicts more explicit and thus more readily detectable. A closer
analysis reveals that this detection is particularly effective on the two
interpretation-conflict types, since these depend on the refined
architecture, and is perfect on discovery conflicts, which are
established deterministically by the matching algorithm. Self-Refine
attains the highest precision but collapses in recall, as its self-critique
loop retains only high-certainty conflicts, whereas RAG reaches a high recall
at the lowest precision, since grounding in textually similar offers rather
than in constraint satisfiability yields spurious conflicts. This advantage carries over to
relaxation: by delegating relaxation to a dedicated module that treats
each conflict independently, our framework achieves the best Recall@5
(83.3\%), against at most 73.2\% for the baselines. The gap is most
pronounced on service discovery, where our deterministic matching attains
the highest precision (98.5\%), recall (84.8\%), and F1 (91.1\%) with no
hallucination. Because the algorithm grounds every candidate in the
actual catalogs, the residual gap to a perfect F1 does not originate
from the matching itself but is inherited from the interpretation phase,
which is carried out by an LLM and therefore remains imperfect. The
generation-based baselines, in contrast, rank offers by proximity and
must trade recall for reliability, recovering barely 60\% of the
compliant offers and hallucinating up to 15.4\% (Mistral-Large-2512,
Self-Refine). RAG is the only baseline that avoids hallucination
(0.0\%), as it selects offers among those retrieved from the catalogs
rather than generating them freely, yet it still forgoes both recall and
precision.\looseness=-1

\subsubsection{Cost analysis.}
\begin{figure}[ht]
    \centering
    \includegraphics[width=0.77\textwidth]{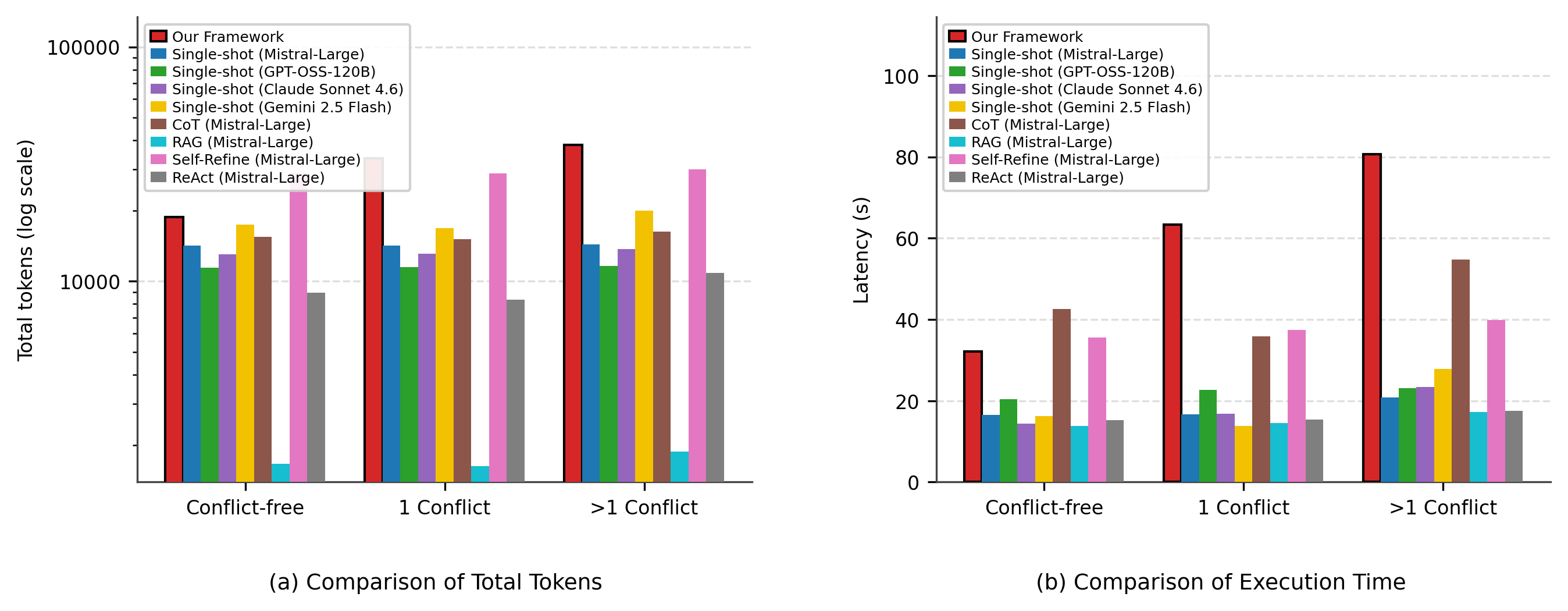}
    \caption{Computational cost comparison across baselines.}
    \label{fig:cost_analysis}
\end{figure}

Fig.~\ref{fig:cost_analysis} reports the token consumption and execution time of our framework compared to the baselines, broken down by the number of conflicts in the request. On conflict-free requests it consumes 18{,}745 tokens and 32.2~s on average, comparable to the Gemini single-shot baseline and below CoT and Self-Refine. Cost grows with the number of conflicts, reaching 47{,}294 tokens and 133.6~s for two-conflict requests, reflecting the extra relaxation and re-matching iterations each conflict triggers.  In contrast, the single-shot baselines exhibit nearly constant costs across different numbers of conflicts, as they perform poorly at detecting and relaxing conflicts, triggering few or no additional iterations.

\begin{table}[htbp]
\centering
\caption{Quality evaluation (Values in \%). Lower block: ablation. Prec.: Precision, Rec.: Recall, R@5: Relaxation Recall@5, Hall.: Hallucination.}
\label{tab:results}
\renewcommand{\arraystretch}{0.82}
\scriptsize
\setlength{\tabcolsep}{3pt}
\begin{tabular}{lcccccccc}
\toprule
\textbf{Method} & \multicolumn{3}{c}{\textbf{Conflict Detection}} & \textbf{Relax.} & \multicolumn{4}{c}{\textbf{Service Discovery}} \\
\cmidrule(lr){2-4} \cmidrule(lr){6-9}
 & \textbf{Prec.} & \textbf{Rec.} & \textbf{F1} & \textbf{R@5} & \textbf{Prec.} & \textbf{Rec.} & \textbf{F1} & \textbf{Hall.} \\
\midrule
Claude Sonnet 4.6 & 59.0 & 85.2 & 69.7 & 70.5 & 92.0 & 60.6 & 73.1 & 0.1 \\
Gemini 2.5 Flash & 62.1 & 66.7 & 64.3 & 68.0 & 87.5 & 53.8 & 66.6 & 7.7 \\
GPT-OSS-120B & 72.7 & 59.3 & 65.3 & 56.1 & 85.7 & 46.2 & 60.0 & 7.7 \\
Mistral-Large-2512 & 58.8 & 74.1 & 65.6 & 73.2 & 83.3 & 53.8 & 65.4 & 15.4 \\
Chain-of-Thought & 63.6 & 77.8 & 70.0 & 72.9 & 91.7 & 59.2 & 72.0 & 7.7 \\
Self-Refine & 88.9 & 29.6 & 44.4 & 36.6 & 84.6 & 51.5 & 64.0 & 15.4 \\
RAG & 40.0 & 74.1 & 51.9 & 70.9 & 90.3 & 59.2 & 71.5 & 0.0 \\
ReAct & 59.4 & 70.4 & 64.4 & 70.7 & 91.2 & 53.8 & 67.7 & 2.3 \\
\midrule
\textbf{Our approach} & \textbf{70.8} & \textbf{89.3} & \textbf{79.0} & \textbf{83.3} & \textbf{98.5} & \textbf{84.8} & \textbf{91.1} & \textbf{0.0} \\
\midrule
\multicolumn{9}{l}{\textit{Ablation:}}\\
w/o Refine. & 40.4 & 84.3 & 54.6 & 75.0 & 98.2 & 38.9 & 55.7 & 0.0 \\
LLM discovery & -- & -- & -- & -- & 94.4 & 59.0 & 72.6 & 5.6 \\
\bottomrule
\end{tabular}
\end{table}
\subsection{Ablation study} 
We further isolate the contribution of two core design choices through dedicated experiments.
\textbf{w/o Refinement.} We remove the reformulation and completion tasks from the interpretation prompt, so that conflicts are detected directly from the raw request, without any intermediate refinement. This variant is evaluated on the full 70-request dataset, across all quality metric dimensions defined above, in order to capture how the absence of request refinement propagates through the rest of the pipeline.
\textbf{LLM-based discovery.} We replace the matching algorithm used for service discovery with an LLM (Mistral-Large-2512), provided with the same catalogs in its prompt and tasked with directly selecting compliant offers. This variant tests whether an LLM can reliably substitute for the deterministic matching logic. Since this change affects only the service discovery phase, it is evaluated on the full dataset, restricted to the service discovery quality metrics (Table~\ref{tab:results}, lower block).
Removing request refinement sharply reduces conflict-detection precision (70.8\%$\to$40.4\%): lacking a completed architecture, the module mistakes missing elements for genuine conflicts. This propagates to discovery, where recall collapses (84.8\%$\to$38.9\%), confirming that refinement is essential for incomplete, high-level requests, without which the requirements passed to discovery stay under-specified. Similarly, replacing the deterministic matching algorithm with an LLM degrades all discovery metrics and, critically, reintroduces a 5.6\% hallucination rate, whereas our deterministic algorithm produces none. These results confirm that service discovery is an information-retrieval task requiring exact, binary decisions over properties and policies, as the LLM omits a substantial share of available compliant offers and occasionally proposes non-existent services.

\section{Discussion and Conclusion}
\label{sec:conclusion}
In this paper, we present a negotiation-aware framework for natural-language request interpretation and service discovery in cloud service composition. It refines high-level, incomplete user intents, dynamically detects multi-type conflicts, generates provider-independent TOSCA templates, and uses a deterministic matching algorithm to retrieve compliant service offerings. Any remaining conflicts in interpretation or unsatisfiable requirements identified during discovery are resolved through an interactive, LLM-based natural-language negotiation loop with the user. Our evaluation demonstrates that the proposed approach significantly outperforms baseline methods, achieving superior precision, recall, and hallucination-free retrieval across a variety of request types. The ablation study further highlights the contribution of the initial request refinement phase and the deterministic discovery algorithm.

\textbf{Threats to validity.} Four limitations temper these results.
Our evaluation runs in fully automatic mode, where the system retains the
minimal proposed relaxation rather than negotiating with a real user, and
therefore does not assess the clarity of the proposed relaxations or user
satisfaction. Our dataset of 70 requests, part of them authored around specific
conflicts, remains small and does not cover a large range of conflict types\MS{; this construction bias is mitigated by writing each conflict in three request forms and pairing it with a conflict-free near-miss request}.
Discovery is evaluated on static catalogs of 209 offers from three providers,
which may not reflect production scale and heterogeneity. Finally,
interpretation, relaxation, and structuring depend on a specific LLM, and their
quality may vary with the model\MS{; moreover, results come from a single run, without variance estimates}. Provider-independent representations and deterministic matching mitigate these threats, and the
directions below address the rest.\looseness=-1 

\textbf{Future work.} We plan to validate the negotiation with real
users, assessing the clarity of the proposed relaxations and user satisfaction\MS{, and to reduce negotiation turns by dynamically adjusting the remaining options, with backtracking, once the user selects one};
to move from static catalogs to real-time discovery over live provider APIs; and
to extend the framework to service selection via multi-criteria optimization\MS{, covering cross-service compatibility,} and,
ultimately, to the cloud--edge--IoT continuum.
\vspace{-5pt}

\begin{credits}
\subsubsection{\ackname} This work was supported by French government grants managed by the Agence Nationale de la Recherche under the France 2030 program, under references ANR-23-PECL-0008 and ANR-22-PEFT-0002.
\end{credits}
%
%
\bibliographystyle{IEEEtran}
\bibliography{reference}

\end{document}